\documentclass[10pt,twocolumn,letterpaper]{article}

\usepackage{cvpr}
\usepackage{times}
\usepackage{epsfig}
\usepackage{graphicx}
\usepackage{amsmath}
\usepackage{amssymb}
\usepackage{amsfonts} 
\usepackage{multirow}

\usepackage[pagebackref=true,breaklinks=true,letterpaper=true,colorlinks,bookmarks=false]{hyperref}

\cvprfinalcopy 

\def\cvprPaperID{2517} 

\ifcvprfinal\fi
\begin{document}

\title{Action Recognition via Pose-Based Graph Convolutional Networks with Intermediate Dense Supervision}

\author{
	Lei Shi$^{1,2}$ \and Yifan Zhang$^{1,2}$\thanks{Corresponding Author} \and Jian Cheng$^{1,2,3}$ \and Hanqing Lu$^{1,2}$ \and
	$^1$National Laboratory of Pattern Recognition, Institute of Automation, Chinese Academy of Sciences\\
	$^2$University of Chinese Academy of Sciences\\
	$^3$CAS Center for Excellence in Brain Science and Intelligence Technology\\
	{\tt\small \{lei.shi, yfzhang, jcheng, luhq\}@nlpr.ia.ac.cn} 
}

\maketitle

\begin{abstract}

Pose-based action recognition has drawn considerable attention recently. 
Existing methods exploit the joint positions to extract the body-part features from the activation map of the convolutional networks to assist human action recognition. 
However, these features are simply concatenated or max-pooled in previous works. The structured correlations among the body parts, which are essential for understanding complex human actions, are not fully exploited. 
To address the problem, we propose a pose-based graph convolutional network (PGCN), which encodes the body-part features into a human-based spatiotemporal graph, and explicitly models their correlations with a novel light-weight adaptive graph convolutional module to produce a highly discriminative representation for human action recognition.
Besides, we discover that the backbone network tends to identify patterns from the most discriminative areas of the input regardless of the others. 
Thus the features pooled by the joint positions from other areas are less informative, which consequently hampers the performance of the followed aggregation process for recognizing actions. 
To alleviate this issue, we introduce a simple intermediate dense supervision mechanism for the backbone network, which adequately address the problem with no extra computation cost during inference. 
We evaluate the proposed approach on three popular benchmarks for pose-based action recognition tasks, i.e., Sub-JHMDB, PennAction and NTU-RGBD, where our approach significantly outperforms state-of-the-arts without the bells and whistles.
\end{abstract}

\begin{figure}[!htb]
\centering
\includegraphics[width=0.95\columnwidth]{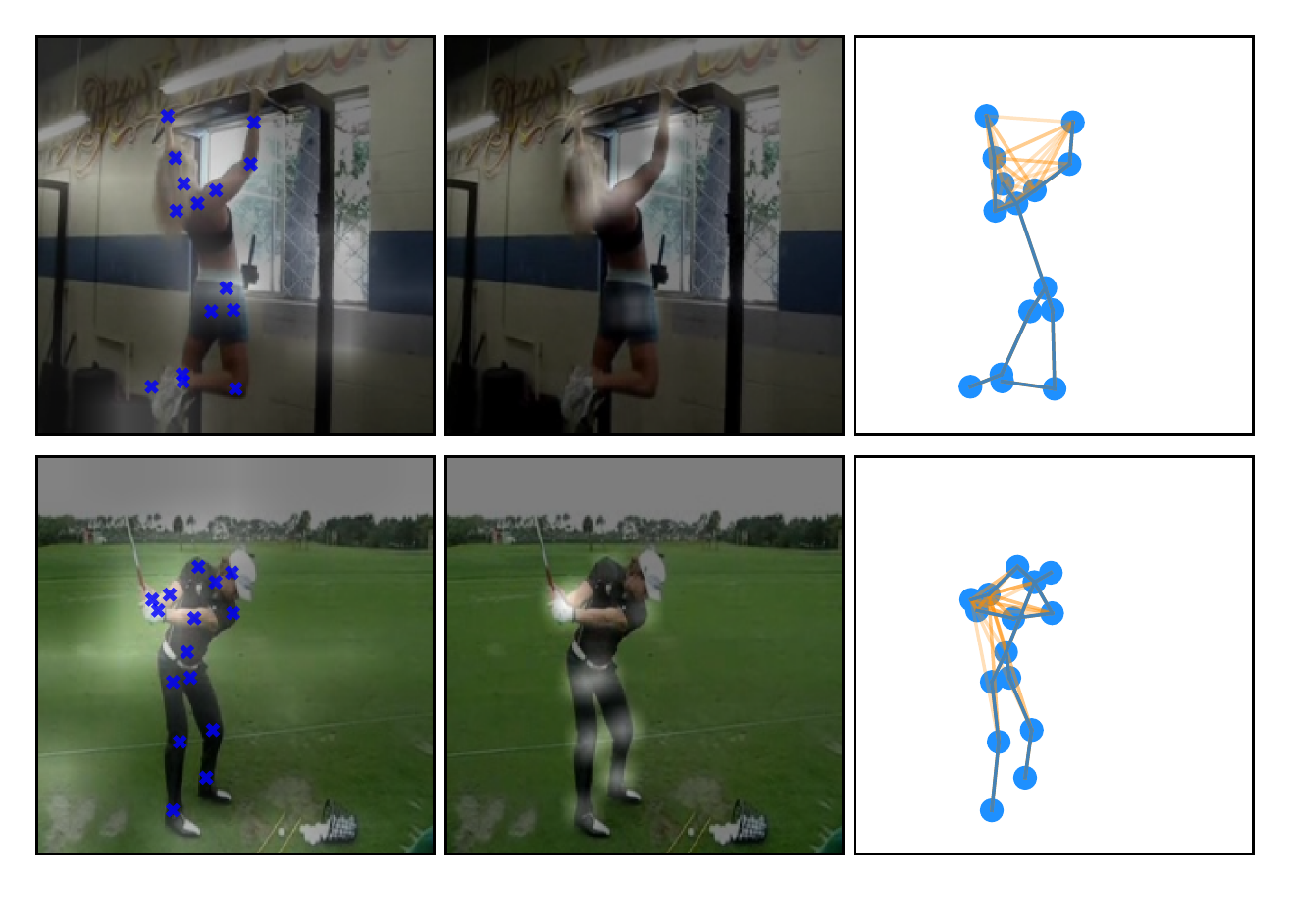} 
\caption{Visualization for actions of the pull-up (the first row) and swing-baseball (the second row) learned by PGCN. 
The first column shows the frames, joints (marked as blue x) and the corresponding activated areas (the brighter areas) in I3D's activations. 
The second column shows the frame and the corresponding activated areas in PGCN's feature map. 
Note that the PGCN focuses more on the human joints that are highly discriminative for the actions, e.g., the arms for pull-up and hands for swing-baseball.
The third column shows the learned graph topology of the PGCN for the two samples. 
The blue circles and lines form the intrinsic graph represent the physical structure of the human body, and the orange lines represent the new-learned edges, which reflect the correlations among the human joints learned by the PGCN for the target actions. 
}
\label{fig:vis}
\end{figure}

\section{Introduction}
Human action recognition has been 
studied for decades since it 
can be widely used in a number of applications such as human-computer interaction and intelligent surveillance~\cite{soomro_ucf101:_2012,cheron_p-cnn:_2015,kay_kinetics_2017,xie_rethinking_2018,shi_gesture_2019,yan_pa3d:_2019}. 
However, human action recognition from the wild videos is still very challenging, mainly due to the high dimension of the video data and the interference of the cluttered background. 
Currently, the mainstream methods for video-based action recognition are built upon two types of inputs, i.e., the RGB images and the corresponding optical flow fields~\cite{wang_temporal_2018,carreira_quo_2017}. 
However, these methods generally use the whole image as input, which may result in that the model does not focus on the human body or specific body parts. The model may learn features from background area due to the training data bias.

Recently, several pose-guided approaches for human action recognition have been proposed to address the 
problem~\cite{cheron_p-cnn:_2015,cao_action_2016,du_rpan:_2017,huang_part-aligned_2019}. 
They employ pose information to explicitly pool the body-part features from the final feature maps of the classification network. Then the extracted features are aggregated for action classification. This can be considered as a regularization mechanism by pose information to make the model focus on the human body.
However, in existing methods, the features pooled from different body positions are independently used or simply concatenated or averaged to obtain a representation. 
Concatenating or averaging the body-part features can not well model the structured dependency and correlation of the body parts as the correlated and uncorrelated body parts are given equal weights.


To move beyond such limitations, we propose a novel pose-based graph convolutional network (PGCN) for human action recognition. 
Instead of simply concatenating or averaging the pooled body-part features, 
we encode them into a spatiotemporal graph constructed based on the physical structure of the human body, and explicitly model the body-related correlations among these features with a novel light-weight adaptive graph convolutional module.
By taking into account these correlations and dependencies, our model can produce a more discriminative pose-related representation for understanding human actions compared with previous methods. 

Another notable problem we discover is that the backbone CNN, from which the body-part features are extracted, is ``lazy".
It is inclined to identify patterns from the most discriminative areas of the inputs, regardless of the others. 
For instance, given a video containing the pull-up action, the 
CNN can 
recognize it by identifying the posture of the arms regardless of the remaining parts such as the hip and legs. 
It is because for most of the popular classification networks, the last feature 
map is global-average-pooled and supervised with only one cross-entropy loss. 
Thus once a part feature is discriminative enough to obtain a high accuracy for the training set, other parts can not generate useful losses anymore. 
It causes the problem that the network makes little effort to identify the patterns of 
those less-discriminative areas. When pooling features from those areas, it is hard for the followed aggregation module to model the body part correlations with these less informative features.

To tackle this problem, we propose to add a novel intermediate dense supervision mechanism for the backbone network during the training process. 
The main idea is to encourage the network to infer the categories using only part of the input information. 
In detail, we drop the global-average pooling layer and directly supervise all elements of the last layer of feature maps of the backbone network with a dense-head classification loss.
This method is simple but effective, without the need for extra parameters and computations during inference. It forces the network to look beyond the most discriminative parts and mine 
rich cues from all areas of the input. 
In this way, the extracted body-part features that are less informative in previous methods can contain more abundant information for inferring the categories, and the followed graph convolutional module can better model the body-related correlations to recognize human actions. 


To test the effectiveness of the proposed methods, we perform extensive experiments on three widely used datasets for pose-based action recognition, namely, Sub-JHMDB, PennaAction and NTU-RGBD, where our model achieves state-of-the-art performance on both of them without the bells and whistles. 
Note that, except for the RGB and pose modalities, we do not use additional modalities such as the optical flow fields and the part affinity fields. 

Overall, our contributions lie in three aspects: 
    \textbf{Firstly}, we propose a pose-based graph convolutional network (PGCN), which can 
    employ the joint position
    to pool the body-part features and aggregate them to better represent human actions. 
    It is the 
    first to employ the graph convolutional module to model the spatiotemporal correlations among the pooled features to produce a highly discriminative pose-related representation for human action recognition. 
    \textbf{Secondly}, we discover that the backbone network tends to ignore identifying patterns from the less discriminative areas of the input in existing methods, which hampers the performance of the followed aggregation process. We propose to add a novel intermediate dense supervision after the backbone network, which effectively addresses the problem without the need for extra parameters and computations. 
    \textbf{Thirdly}, we evaluate our approach on three popular benchmarks for pose-based action recognition, which achieves state-of-the-art performance using only the RGB and pose modalities without additional training tricks. Our code is released to provide a new baseline and facilitate communication. 

\begin{figure*}[htb]
\centering
\includegraphics[width=0.9\textwidth]{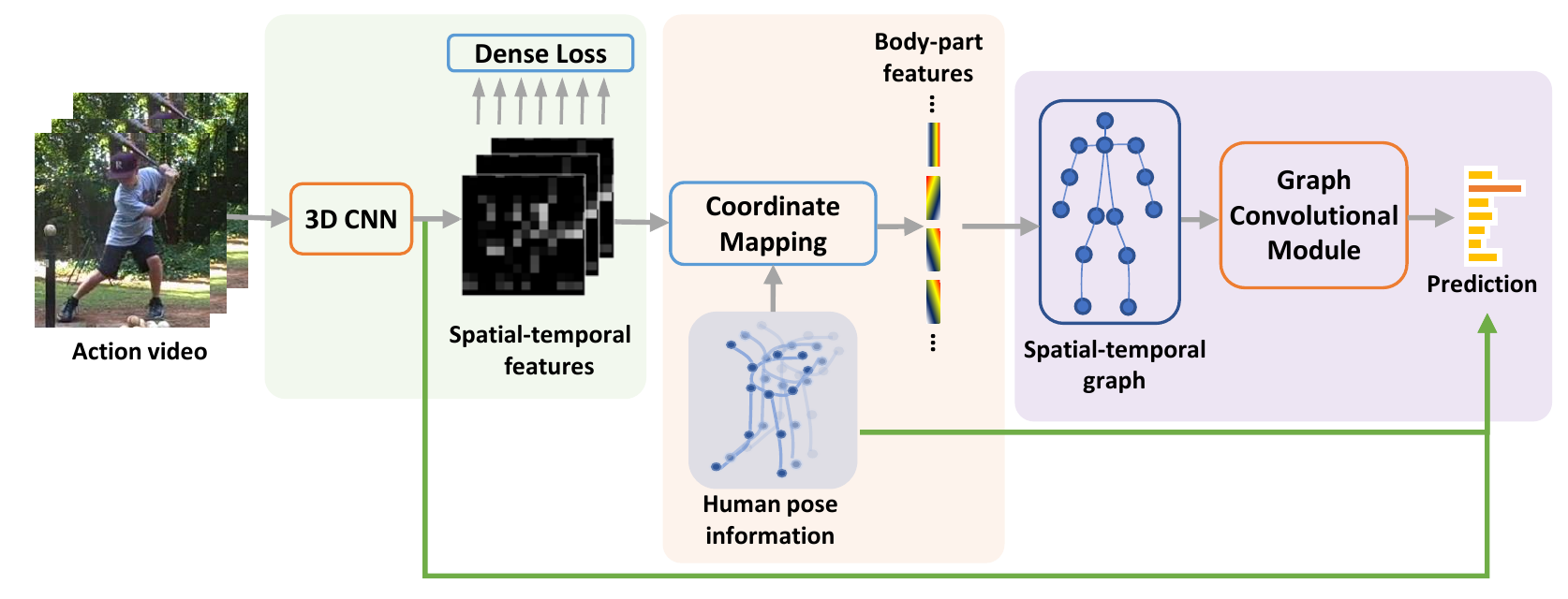} 
\caption{The pipeline of the proposed PGCN. }
\label{fig:pipeline}
\end{figure*}

\section{Related Works}
\label{sec:relatedwork}
\subsection{Action recognition}
\label{sec:re-action}
Early approaches for action recognition focus on designing hand-crafted features~\cite{wang_action_2011,wang_action_2013}. 
Recently, deep neural networks have shown big success in many computer vision tasks, which also exhibit excellent performance in the field of action recognition. 
There mainly exists three branches for these data-driven approaches: RNN-based~\cite{yue-hei_ng_beyond_2015,ma_long-term_2015} approaches, two-stream~\cite{simonyan_two-stream_2014,wang_temporal_2018} approaches and 3D-CNN-based approaches~\cite{tran_learning_2015,hara_can_2018}.
Among these approaches, the 3D-CNN-based approaches usually achieve higher performance and faster speed. 
It can also be integrated with the optical flow to form the two-stream 3D CNNs \cite{carreira_quo_2017,wang_non-local_2018}. 
However, since the action is human-centered, these methods ignore the important cues for recognizing human actions, i.e., the spatiotemporal evolutions of the human pose.
Our work is based on the 3D CNNs, but further incorporates the pose information to help model understanding the complex human actions.

\subsection{Pose-based action recognition}
Human pose has been proved a discriminative cue in previous hand-crafted-feature-based methods for human action recognition~\cite{jhuang_towards_2013,iqbal_pose_2017}. 
Recently, with the emergence of the data-driven methods, several works propose to exploit the human pose in deep models to help classifying human actions ~\cite{cheron_p-cnn:_2015,cao_action_2016,du_rpan:_2017,choutas_potion:_2018,luvizon_2d/3d_2018,liu_recognizing_2018,huang_part-aligned_2019,yan_pa3d:_2019}.
One stream of these methods try to pool a number of body-part features guided with the pose information, which are then aggregated to produce a discriminative representation to predict human actions. 
In detail, Cheron et al.~\cite{cheron_p-cnn:_2015} employ a shared CNN to extract the body-part features from the cropped image patches and aggregate them with the max-min pooling strategy. 
Cao et al.~\cite{cao_action_2016} pool the body-part features from the C3D's feature maps and concatenate them to feed into the SVM classifier. 
Du et al.~\cite{du_rpan:_2017} and Huang et al.~\cite{huang_part-aligned_2019} also pool the joint-related features but from 2D CNNs. Du et al.~\cite{du_rpan:_2017} aggregate the features with a body-part-pooling layer and fed them into an LSTM to model the temporal evolutions.  
Huang et al.~\cite{huang_part-aligned_2019} further propose a part-based hierarchical pooling approach that can group and pool body-part features from local to global. 
However, the body-part features are highly coupled with each other based on the human body structure, and these approaches fail to model the body-related correlations using the straightforward aggregation strategies as introduced above. 
Another stream tries to exploit the by-products of the pose estimation to assist the human action recognition, such as the part affinity fields and the pose estimation heatmap~\cite{liu_recognizing_2018,luvizon_2d/3d_2018,yan_pa3d:_2019}. 
In this work, we do not investigate these by-products since they are not our focus, but it is simple and straightforward to integrate them within a multi-stream framework. 

\subsection{Skeleton-based action recognition}
Another stream, namely, skeleton-based action recognition, use the pure pose information, i.e., the 2D/3D coordinates of the human joints, to recognize human actions~\cite{du_hierarchical_2015,cao_skeleton-based_2018,yan_spatial_2018,shi_two-stream_2019,shi_skeleton-based_2019}. 
However, for these approaches, it is hard to classify the actions that have similar postures such as ``torch chest'' versus ``torch back'' because the appearance information is discarded. 
In our methods, both the joints coordinates and the appearance information are leveraged to solve the problem effectively. 

\subsection{Graph convolutional networks}
Graph convolutional networks (GCNs) can generalize CNNs from grid inputs such as images and videos to non-Euclidean structured data such as graphs and manifolds~\cite{shuman_emerging_2013,defferrard_convolutional_2016,wang_videos_2018,yan_spatial_2018,shi_skeleton-based_2019}.
There are two streams for constructing GCNs: 
1) the spectral-stream approaches formulate the GCNs in the spectral domain, which are based on the analogy between the classical Fourier transforms and projections onto the eigenvectors of the graph Laplacian operator~\cite{shuman_emerging_2013,defferrard_convolutional_2016};
2) the spatial-stream approaches provide filter localization via the finite size of the kernel, which is application-dependent and needs to design the mapping functions manually to match the local neighbors~\cite{wang_videos_2018,yan_spatial_2018,shi_two-stream_2019}. 
Our work follows the spatial-stream approaches, which are more flexible and more targeted for particular tasks.

\section{Our Methods}
In this section, we introduce the proposed pose-guided graph convolutional network (PGCN) in detail. 
First, the overall pipeline of the proposed framework is presented briefly in Sec.~\ref{sec:pipeline}. 
Then, we show how to obtain the body-part features from the 3D CNN's feature maps with the help of the pose information in Sec.~\ref{sec:featureextraction}. 
Besides, the graph convolutional module and some specific designs of the network architecture are described thoroughly in Sec.~\ref{sec:graphconvolutionalmodule}. 
Finally, we introduce the ``laziness" problem of the backbone network and the proposed intermediate dense supervision mechanism in Sec.~\ref{sec:intersuper}.

\subsection{Pipeline overview}
\label{sec:pipeline}
Fig.~\ref{fig:pipeline} shows the overall pipeline of the proposed PGCN.
It is input with the RGB video and the corresponding human pose information, which can be easily obtained from the motion-capture device such as Kinect or the pose estimation algorithm such as OpenPose~\cite{cao_realtime_2017}. 
\textbf{First}, the input video clip is fed into a 3D CNN to extract the spatiotemporal features from low-levels to high-levels. 
To solve the ``laziness" problem of the backbone network, we add a dense-head classification loss after the backbone for intermediate supervision. 
\textbf{Then}, with the help of the coordinate mapping strategy, the body-part features are pooled from the last feature maps of the 3D CNN. 
Owing to the conception of the receptive field~\cite{cao_action_2016,du_rpan:_2017}, these features represent the surrounding information of the corresponding joints in the original video. 
\textbf{Finally}, a spatiotemporal graph is built upon the structure of the human body to encode these features and their correlations, which is fed into a graph convolutional module to produce a pose-related representation and generate the human-based classification scores. 
\textbf{Moreover}, the classification score of the original 3D CNN, which is global-average-pooled from the spatiotemporal feature, naturally represent the information of the whole scenes. 
The skeleton information, i.e., the coordinates of the joint, naturally represents the posture information of the human body. 
We fuse the human-based classification score, the scene-based classification score and the posture-based classification score to get the final prediction (green lines in Fig.~\ref{fig:pipeline}). 
In the rest of the paper, for a fair comparison, we call the mainstream, i.e., without exploiting the joint coordinates and scene-based information, as the PGCN and the overall framework as the PGCN-Fusion. 

\subsection{Feature Extraction}
\label{sec:featureextraction}
In this section, we introduce the feature extraction part of the PGCN, i.e., how to obtain the body-part features. 
As introduced in Sec.~\ref{sec:re-action}, the two-stream 3D CNNs have been proved useful in learning spatiotemporal representations of videos.
We use the spatial stream of the well-known two-stream I3D~\cite{carreira_quo_2017} as the feature extractor. 
We do not use the temporal-stream in this work because calculating the optical flow field is time-consuming. 
It is also worth to note that our approach is not restricted to the 3D CNNs. 
We have also tried other feature extractors such as 2D CNNs~\cite{wang_temporal_2018}, but the performance is not as good as using 3D CNNs. 



For each step of the convolution, the convolutional filter only acts on a local region,
which results in a feature that corresponds to a particular sensory space of the feature map in the preceding layer. 
This local space is called the receptive field of the point outputted in this convolutional step~\cite{cao_action_2016,du_rpan:_2017}. 
Because the 3D CNN employs the 3D filters to extract the spatiotemporal features, its receptive field is a spatiotemporal cube. 
The points in the higher layers of 3D CNNs can be mapped to the corresponding cubes in the lower layers and vice versa. 
By mapping the point layer by layer, we can localize the point in arbitrary feature maps that corresponds to the human joint of the input video. 
Because the estimated coordinates of the human joints may have errors, the features of the $3\times 3\times 3$ cube around the joints in the feature maps are extracted and max-pooled to reduce the wrong mappings. 
We also tried using average-pooling and max-min-pooling for the cube, but the performance is not as good as using the max-pooling. 
Note that there may be some joints that are not detected or outside the image. 
We pad the features of these joints with $0$. 
Another strategy is to associate the semantically-related human joints into a number of groups and replace the missed joints with the one in the same group~\cite{du_rpan:_2017}. 
However, we found it has no effect. 

\subsection{Adaptive graph convolutional module}
\label{sec:graphconvolutionalmodule}

\begin{figure}[htb]
\centering
\includegraphics[width=0.9\columnwidth]{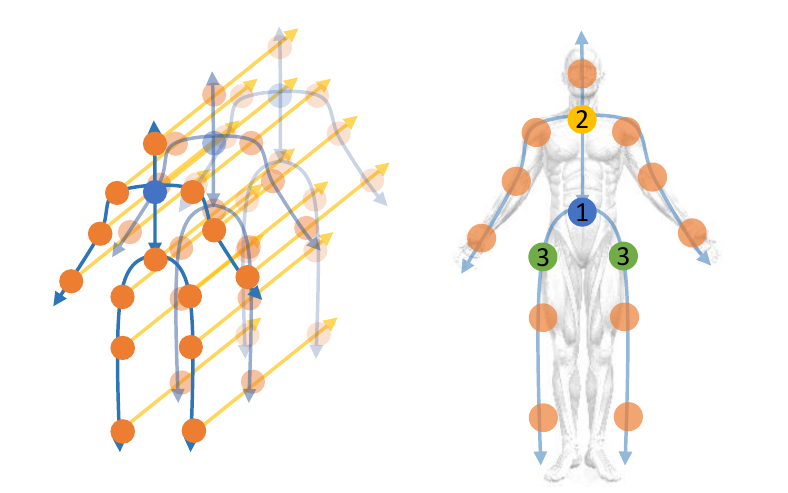}
\caption{Illustration of the spatiotemporal graph building (left) and the subset-partition strategy (right). }
\label{fig:graph}
\end{figure}

\subsubsection{Graph building}
Inspired by previous works for human parsing~\cite{yang_articulated_2011}, we represent the human body as a tree-structured graph, where the nodes are the human joints and the edges are the physical connections among the joints (Fig.~\ref{fig:graph}, left). 
The root node is defined as the gravity center of the human body (the blue node). 
The spatial edges are pointed from the parent nodes to the child nodes (the blue lines). 
The temporal edges are pointed from the joint in the preceding frame to the same joint in the later frame (the yellow lines). 
The attribute of each node is the body-part features introduced in previous sections.  
In this way, a video can be encoded into an acyclic spatiotemporal directed graph. 

\subsubsection{Convolution on graphs}
Performing convolution on graphs is not as straightforward as performing convolution on images. 
It is because the number of weights of a convolutional kernel is fixed, but the number of the node neighbors in the graph is unfixed. 
This leads to the mapping problem between the unfix-numbered neighbor-nodes and the fix-numbered weights. 
Inspired by previous works~\cite{yan_spatial_2018,shi_two-stream_2019}, we solve this problem by partition the neighborhood nodes into fix-numbered subsets, where the number is equal with the kernel size. 

Specifically, given the feature of the node $j$ in the $i_{th}$ feature map and $t_{th}$ frame, i.e., $f^i(v_{tj})$, the corresponding activation in the $(i+1)_{th}$ feature map, i.e., $f^{i+1}(v_{tj})$, is calculated as
\begin{equation}
    f^{i+1}(v_{tj}) = \sum_k^K \frac{w_k}{N_{S_k(v_{tj})}} \sum_{v_{tl} \in S_k(v_{tj})} f^i(v_{tl})
\label{eq:partition}
\end{equation}{}
where $K$ is the kernel size of the convolution. $w_k$ is the weight of the kernel. $S_k(v_{tj})$ denotes the node subset of $v_{tj}$ that corresponds to the $w_k$. 
$N_S$ denotes the number of the nodes in the subset $S$. 
The nodes in the subset $S_k(v_{tj})$ will be average-pooled and multiplied with $w_k$. 
Because the number of the subsets is the same with the kernel size, the mapping problem is avoided. 
Note that if $N_{S_k(v_{tj})}=1$, Eq.~\ref{eq:partition} is  same with the standard convolution. 

Now, the main problem lies in how to partition the subsets. 
For the spatial dimension, the neighborhood of one node is defined as the 1-distance nodes in the graph. 
Since the graph is tree-structured, we set the kernel size to three and divide the 1-distance neighborhood of one node into three subsets: the subset of its parent nodes, the subset of itself and the subset of its child nodes. 
This representation is intuitive because the human body is naturally an articulated system, where the position of one joint is always affected by its parent joint and affects the position of its child joints. 
Fig.~\ref{fig:graph} (right) shows this partitioning strategy, where the numbered circles are the neighbors of the node $1$ and different numbers represent different subsets. 
For the temporal dimension, since the temporal order is fixed, one node must have two neighbors, i.e., the same nodes in the preceding frame and the next frame. 

\subsubsection{Implementation}
\label{sec:implementation}
Implementation of the convolution on graphs is not directed, especially for the spatial dimension where the nodes have no implicit arrangement. 
Here, we exploit the adjacency matrix of the graph to achieve the operation. 
Given the extracted body-part features in the tensor form $\mathbf{f}_{in}\in\mathbb{R}^{C_{in}\times T\times J}$ where $T$ and $J$ denote the number of frames and the number of joints, the convolutions on graph can be formulated as
\begin{equation}
    \mathbf{f}_{out} = \sum_k^{K_s} \mathbf{W}_k (\mathbf{f}_{in} \mathbf{\hat{A}}_k)
    \label{eq:implement}
\end{equation}
where $K_s$ denotes the kernel size along the spatial dimension. $\mathbf{W}_k \in\mathbb{R}^{C_{out}\times C_{in}}$ is the weight matrix. 
$\mathbf{\hat{A}}_k = \mathbf{\Lambda}_k^{-\frac{1}{2}} \mathbf{A}$. 
$\mathbf{A}_k \in\mathbb{R}^{J\times J}$ is an adjacency-like matrix that denotes the connectivity of the subset $S_k$. 
Its element $\mathbf{A}_k^{xy}$ denotes whether the node $v_x$ is in the subset of the node $v_y$, i.e., $S_k(v_y)$.
$\mathbf{\Lambda}_k^{-\frac{1}{2}}$ is a diagonal matrix that is used for normalization, i.e., averaging the nodes in the subset. 
Its element $\mathbf{\Lambda}_k^{xx} = \sum_y \Lambda_k^{xy}$.
As for the temporal dimension, the graph convolution is equipped with performing a 2D convolution with the kernel size as $(K_t, 1)$ on the input tensor  $\mathbf{f}_{in}\in\mathbb{R}^{C_{in}\times T\times J}$. 
Where $K_t$ denotes the kernel size along the temporal dimension. 

\subsubsection{Adaptive graph}
\label{sec:adaptivegraph}
Inspired by \cite{shi_two-stream_2019,wang_videos_2018}, we adapt the graph topology.
The original graph is defined according to the physical structure of the human body, which may not be the best choice for the action recognition task. 
For example, when recognizing the action of ``clapping", the correlations between two hands should be significant. 
However, they are far from each other in the human-body-based graphs. 
Note that the graph topology is determined by the adjacency matrix as introduced in Sec.~\ref{sec:implementation}.
Here, we add two additional adaptive graphs: the global graph and the individual graph. 
The global graph is the same with the human-body-based graph, but is set as the parameter and is updated based on the action recognition loss in the training process. 
The individual graph is calculated based on the similarities among the body-part features.
Eq.~\ref{eq:ck} shows the calculation of the graph structure. 
\begin{equation}
\begin{aligned}
    \mathbf{A}_k & = \mathbf{B}_k + \alpha \mathbf{C}_k \\
    \mathbf{C}_k & = Tanh(\mathbf{f_{in}}^T\mathbf{W}^T_{\theta k}\mathbf{W}_{\phi k}\mathbf{f_{in}})
\end{aligned}
\label{eq:ck}
\end{equation}
where $\mathbf{A}_k$ is the adjacency-like matrix introduced in Eq.~\ref{eq:implement}. 
$\mathbf{B}_k$ denotes the global graph and $\mathbf{C}_k$ denotes the individual graph. 
$\theta$ and $\phi$ are two embedding functions which are realized with the $1\times 1$ convolutions. 
Note that we use the $Tanh$ instead of the $SoftMax$ to normalize the $\mathbf{C}_k$. 
We argue that the $Tanh$ can provide more flexibility to learn the graph topology.
For example, it can remove the edges learned in the global graph based on the individual requirements of the current data, but the $SoftMax$ can only produce the positive value. 
Besides, $Tanh$ is insensitive to large values and can mine more correlations among the features.

\subsection{Intermediate dense supervision}
\label{sec:intersuper}
\begin{figure}[!htb]
    \centering
    \includegraphics{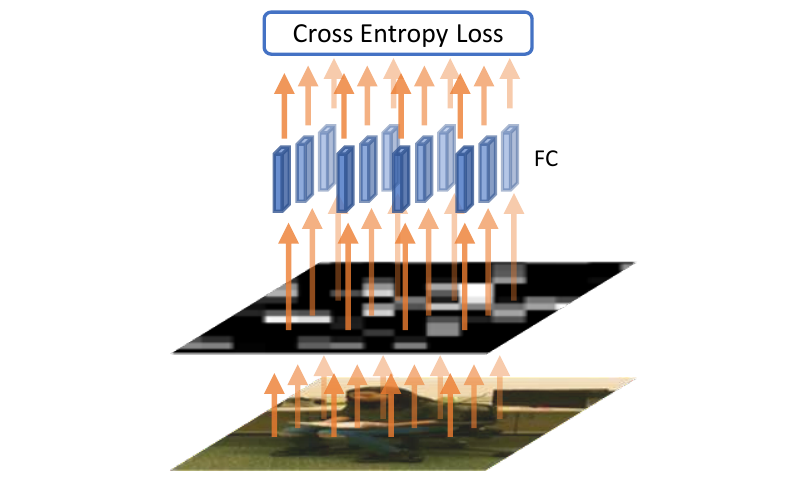}
    \caption{Illustration of the intermediate dense supervision.}
    \label{fig:denseloss}
\end{figure}

The last feature map of the 3D CNN is a spatial-temporal volume  
$\mathcal{V}\in \mathit{R}^{T\times H\times W\times C}$, where $T$, $H$, $W$ and $C$ denote the length, height, width of the feature map and the number of channels, respectively. 
It consists of $T\times H\times W$ $C$-dimension features each corresponds to a particular cube of the original video as illustrated in Section~\ref{sec:featureextraction}. 
To encourage the network to fully extract the information contained in every cube of the input video, we drop the global-average-pooling layer of the I3D and require that every feature of $\mathcal{V}$ should be fully exploited and make a prediction of the actions.  
In detail, as shown in Fig.~\ref{fig:denseloss}, we add $T\times H\times W$ heads after the last feature maps, each of which is realized with a single fully-connect layer. 
Every head inputs the corresponding $C$-dimension feature vector and outputs the label prediction. 
The cross-entropy loss is added after each head and averaged to back-propagate the errors. 
Note that the parameters of the heads are not shared so that the back-propagation process of each branch does not affect each other.
In this way, it forces the model to identify patterns from all areas of the input video rather than only focus on the most discriminative one. 

{}


\section{Experiments}
\subsection{Datasets}
In our experiments, we choose three popular datasets for pose-based action recognition, namely, Sub-JHMDB~\cite{jhuang_towards_2013}, PennaAcion~\cite{zhang_actemes_2013} and NTU-RGBD~\cite{shahroudy_ntu_2016}, to verify the effectiveness of the proposed methods. 
We do not use Kinetics because the humans are poorly visible in many videos of this dataset, which is also discussed in previous related works~\cite{choutas_potion:_2018,yan_pa3d:_2019,huang_part-aligned_2019}. Instead, we use NTU-RGBD as a substitute, which is large, challenging and widely used for pose-based action recognition.

\textbf{Sub-JHMDB} is a subset of JHMDB, which has 316 videos with 12  human action classes. Each video is annotated with the action label and 15 body joints. We conduct the experiments in the three official provided splits of the dataset. 

\textbf{PennaAction} consists of 2325 videos in the wild with 15 action classes. Each video is annotated with 13 body joints. We conduct the experiments with the  50/50 training/testing split provided by the dataset.  Note that there are videos in which not all the body joints are visible. 

\textbf{NTU-RGBD} consists of 56,000 action clips in 60 action classes. There are up to 2 subjects in each video, where each subject is annotated with 25 joints. We use the cross-subject benchmark to evaluate the performance, which means the subjects are different for the training set and test set. Since the person is relatively small compared with the whole scene, we crop the person area based on the pose information. 

\subsection{Training details}
To highlight the keys instead of the training skills, we use a simple strategy to process the input data. 
For training, we randomly sample 40 frames and temporally randomly crop 32 frames. 
Each frame is spatially randomly cropped to $256\times 256$, and resized to $224\times 224$. 
As for the pose information, the positions of the joints are randomly perturbed within $1\%$. 
For testing, we use uniform-sample and center-crop strategies. 

When training the PGCN,  we use the I3D pretrained in the Kinetics~\cite{carreira_quo_2017} and fine-tune it in the target dataset. 
The initial learning rate for fine-tuning I3D is $0.01$, which is divided by 10 after the validation accuracy saturates. Weight decay is $0.0005$. 
We use four TITANXP-GPUs and the batch size is 32. 
Then, we fix the parameters of the backbone model and train the graph convolutional module. 
The learning rate is initialized to $0.1$, and is also divided by 10 after the validation accuracy saturates. 
Weight decay is $0.0001$ and batch size is 32. 
Finally, both the I3D and graph convolutional module are fine-tuned together with the learning rate being $0.001$. 



\subsection{Ablation study}
In this section, we test the effectiveness of the proposed modules on the first split of the Sub-JHMDB since it is relatively smaller. 
When testing one configuration, the others are set the same for a fair comparison. 

\subsubsection{Adaptive graph convolutional module}
We test the necessary of pooling the body-part features and the effectiveness of the proposed graph-based aggregation strategy as shown in Tab.~\ref{tab:posesampling}.
I3D denotes the baseline that does not use the pose information. 
I3D+FC means pooling the body-part features based on the pose information and uses a simple fully-connected layer to aggregate the features. 
PGCN denotes our method that employs the graph convolutional module to model the structured correlations among these body-part features.
It shows that exploiting the pose information (I3D+FC) performs better compared with the baseline method (I3D).
Besides, employing the proposed graph convolutional module (PGCN) further improves the performance with nearly no additional parameters and computations. 
It confirms our motivation that modeling the correlations and dependencies among the body parts is important for understanding human actions. 
Moreover, we also test the performance of using the pose information obtained by the  OpenPose~\cite{cao_realtime_2017} (PGCN*). 
It shows that the performance drops a little since the pose estimated by OpenPose is less accurate than GT-annotations. 

\begin{table}[!htb]
\centering
\begin{tabular}{l|c|c|c}
\hline
Method & Accuracy & Params(M) & GMACS \\
\hline\hline
I3D         & 85.4          & 12.33 & 55.75\\  
I3D+FC      & 86.5          & 13.06 & 55.75\\ 
PGCN        & \textbf{88.8} & 13.07 & 55.78\\ 
PGCN*       & 87.6          & 13.07 & 55.78\\ 
\hline
\end{tabular}
\caption{Recognition accuracy (\%) of different strategies for exploiting the pose information on the spilt $1$ of the Sub-JHMDB. * denotes using the pose information estimated by the OpenPose~\cite{cao_realtime_2017}. }
\label{tab:posesampling}
\end{table}

We also investigate the effectiveness of adaptively learning the graph topology as shown in Tab.~\ref{tab:graph}. 
There are two types of adaptive graphs: the global graph (B) and the individual graph (C). 
It shows that both of the two types of graphs bring consistent improvements.

\begin{table}[!htb]
\centering
\begin{tabular}{l|c}
\hline
Method & Accuracy  \\
\hline\hline
no adaptive & 84.3 \\  
B only & 86.5 \\ 
B+C & \textbf{88.8} \\ 
\hline
\end{tabular}
\caption{Recognition accuracy (\%) of different adaptive graph configurations on the spilt $1$ of the Sub-JHMDB.}
\label{tab:graph}
\end{table}

We visualize two examples for actions of the pull-up (the first row) and the swing-baseball (the second row) in our model in Fig.~\ref{fig:vis}. 
The first column shows the activation areas of the I3D's last convolutional layer in one frame. 
It shows that the activation areas are distributed throughout the scenes.  
The second column shows the activation areas of the PGCN's last layer. 
Because it is input with the joint-related features, the PGCN only cares about the human-centered areas. 
It shows the PGCN focuses more on the particular joints that are discriminative for the actions. 
For example, when modeling the sample of the pull-up, it activates more on the arms of the human. 
When modeling the sample of the swing-baseball, it activates more on the two hands. 
The third column shows the learned graph topology of the adaptive graph convolutional module. 
For pull-up, the new edges are mainly built between the two arms, which means the model successfully models the correlations between the two arms. 
For swing-baseball, the PGCN focuses more on the correlations between the hands and other joints. 
This consolidates the effectiveness of the proposed adaptive graph convolutional module.


\subsubsection{Intermediate dense supervision}

\begin{figure*}[htb]
\centering
\includegraphics[width=0.85\textwidth]{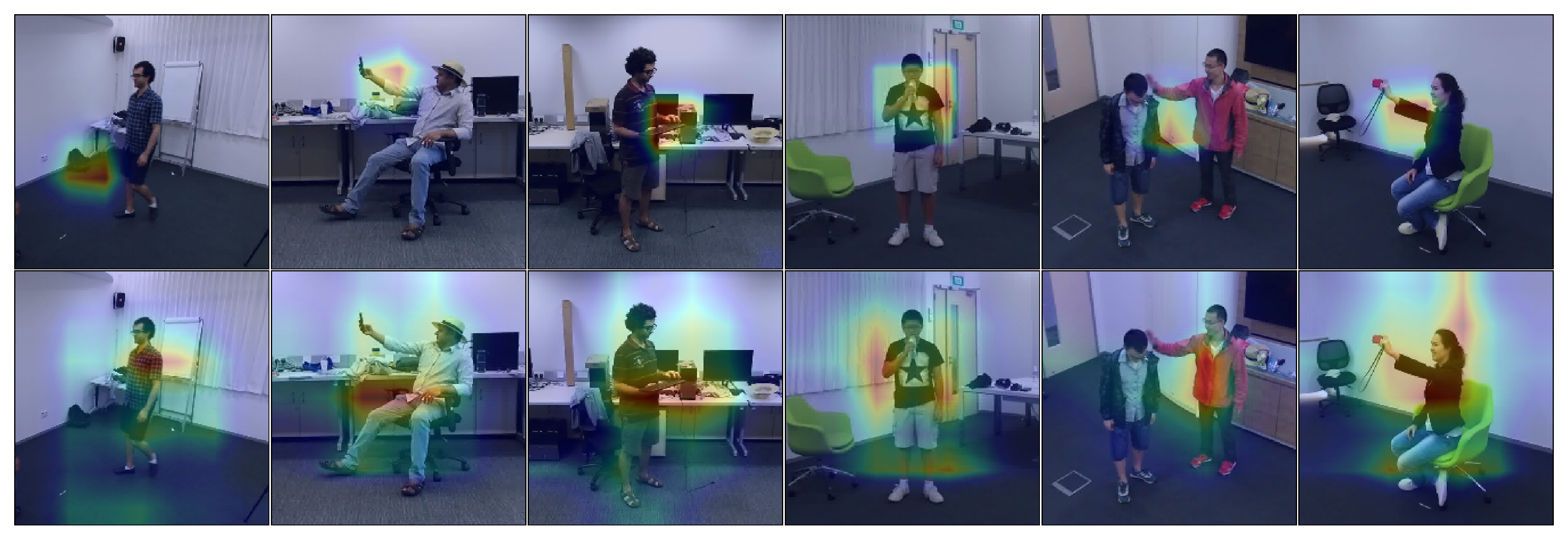} 
\caption{Comparison of the activations of the I3D without (the first row) and with (the second row) the intermediate dense supervision (IDS). 
The I3D supervised by the IDS actives more than original I3D.  
Although the loss makes the model focus on some background areas, the pose-based pooling mechanism and the graph convolutional module can effectively solve this problem.
Note that in the first example, the original I3D mistakenly focuses the bag and totally ignores the people. By adding the IDS, the problem is avoid.  }
\label{fig:feature_diff}
\end{figure*}

We test the effectiveness of the intermediate dense supervision (IDS) introduced in Section~\ref{sec:intersuper}. 
As shown in Tab.~\ref{tab:intersuper}, using the IDS harms the performance of the original 3D CNN (``I3D w/ IDS" vs. ``I3D w/o IDS"). 
It is because most of the areas of the input video do not have enough information to recognize the actions, and averaging their outputs has a negative impact on the final result. 
However, when adding the fully-connected layers or the graph convolutional module, the performance of using the intermediate supervision is better (``PGCN w/ IDS" vs. ``PGCN w/o IDS" and ``I3D+FC w/ IDS" vs. ``I3D+FC w/o IDS"). 
It is because the IDS encourages the model to identify patterns from every area of the inputs rather than only the most discriminative area.
Thus the body-part features that are less informative in previous methods now contain more abundant information, and modeling their correlations brings more improvement. 

Fig.~\ref{fig:feature_diff} shows some examples of the learned feature maps of the ``I3D w/o IDS" (first row) and the ``I3D w/ IDS" (second row). 
It shows the I3D supervised by IDS actives more than original I3D. 
Although the loss makes the model focus on some background areas, the pose-based pooling mechanism and the graph convolutional module can effectively solve this problem.
Note that in the first example, the original I3D mistakenly focuses the bag and totally ignores the people. By adding the IDS, the problem is avoid. 

\begin{table}[!htb]
\centering
\begin{tabular}{c|c|c}
\hline
Model   & w/o IDS &  w/ IDS \\
\hline\hline
I3D     & \textbf{85.4} & 81.5 \\  
I3D+FC  & 86.5 & \textbf{87.6} \\  
PGCN    & 88.8 & \textbf{89.9} \\ 
\hline
\end{tabular}
\caption{Recognition accuracy (\%) on the spilt $1$ of the Sub-JHMDB. IDS denotes the intermediate dense supervision. w/o denotes without and w/ denotes with.}
\label{tab:intersuper}
\end{table}

\subsubsection{Feature fusion}
We test the complementarity among different modalities in Tab.~\ref{tab:fusion}. 
``Coordinate" means feeding only the coordinate of the joints into the graph convolutional module to predict actions. 
Since it lacks the appearance information, its performance is lower. 
By fusing both of the RGB image and the pose information, our PGCN outperforms the single modality based methods.
Furthermore, by fusing them all in a unified framework, the performance is improved significantly. 
\begin{table}[!htb]
\centering
\begin{tabular}{l|c}
\hline
Method & Accuracy \\
\hline\hline
Coordinate & 69.7 \\  
I3D & 85.4 \\ 
PGCN & 89.9 \\ 
PGCN-Fusion & \textbf{91.0} \\
\hline
\end{tabular}
\caption{Recognition accuracy (\%) of using different modalities on the split 1 of the Sub-JHMDB. }
\label{tab:fusion}
\end{table}

\subsection{Comparison with State of the Arts}
In this section, we first compare the recognition accuracy of our PGCN with other approaches on the Sub-JHMDB and PennaAction. 
The results of the Sub-JHMDB are averaged over three official splits. 
Tab.~\ref{tab:sota} shows that our PGCN achieves state-of-the-art performance on both of the two datasets (+2.8\% and +0.8\%) without the bells and whistles. 
Note that ``P-Evo.'' and ``DPI+DTI'' additionally exploit the joint estimation map, i.e., the by-product of the pose estimation algorithm. 
``P$^2$RN'' additionally exploits the optical flow fields. 
Our method only uses the pose information and the RGB videos as input. 
We further evaluate the PGCN on a larger widely used dataset, i.e., NTU-RGBD, as shown in Tab.~\ref{tab:ntu},
where our PGCN outperforms state-of-the-arts significantly (+4.7\%).

\begin{table}[!htb]
\centering
\begin{tabular}{l|c|c|c}
\hline
State-of-the-art & Year & Sub-JHMDB & PennaAction \\
\hline\hline
P-CNN~\cite{cheron_p-cnn:_2015}                 & 2015 & 72.5 & - \\
JDD~\cite{cao_action_2016}                      & 2016 & 83.3 & 95.7\\
RPAN~\cite{du_rpan:_2017}                       & 2017 & 78.6 & 97.4\\
Pose+MD~\cite{huang_human_2018}                 & 2018 & 78.9 & 97.6\\
P-Evo.~\cite{liu_recognizing_2018}              & 2018 & - & 98.2 \\
DPI+DTI~\cite{liu_joint_2019}                   & 2019 & - & 95.9  \\
P$^2$RN~\cite{huang_part-aligned_2019}          & 2019 & 86.5 & 98.0\\
\hline
PGCN                                            & 2019 & 87.0 & 98.1\\
PGCN (Fusion)                                   & 2019 & \textbf{89.3} & \textbf{99.0} \\
\hline
\end{tabular}
\caption{Comparison with state-of-the-arts on the Sub-JHMDB (average over three splits) and PennaAction.}
\label{tab:sota}
\end{table}


\begin{table}[!htb]
\centering
\begin{tabular}{l|c|c}
\hline
State-of-the-art & Year & Accuracy \\
\hline\hline
ST-GCN~\cite{yan_spatial_2018}                  & 2018 & 81.5 \\
DGNN~\cite{shi_skeleton-based_2019}             & 2019 & 89.9 \\
\hline
Luvizon et al.~\cite{luvizon_2d/3d_2018}        & 2018 & 85.5  \\
Baradel et al.~\cite{baradel_glimpse_2018}      & 2018 & 86.6  \\
P-Evo.~\cite{liu_recognizing_2018}              & 2018 & 91.7  \\
DPI+DTI~\cite{liu_joint_2019}                   & 2019 & 90.2  \\
\hline
PGCN                                            & 2019 & 94.0 \\
PGCN (Fusion)                                   & 2019 & \textbf{96.4}  \\
\hline
\end{tabular}
\caption{Comparison with state-of-the-arts on the cross-subject benchmark of the NTU-RGBD dataset.}
\label{tab:ntu}
\end{table}

\section{Conclusion}
In this paper, we propose a novel pose-based graph convolutional network (PGCN) for human action recognition.
Unlike previous approaches for pose-based action recognition, it builds a spatiotemporal graph with the body-part features pooled from the last feature map of the 3D CNN and employs a light-weight adaptive graph convolutional module to model the body-related correlations among these features.
Besides, we point that the backbone network tends to ignore identifying patterns from less discriminative ares in existing approaches, which hampers the performance of the followed aggregation process. 
To address the problem, we introduce a novel intermediate dense supervision mechanism, which is verified effective with no extra parameters and computation cost during inference.
Our method is evaluated on three popular datasets for pose-based action recognition task and outperforms state-of-the-arts significantly on all of them. 
Future work can focus on how to integrate the human pose estimation algorithms and the pose-guide action recognition algorithms in a unified framework.

\newpage
\bibliographystyle{ieee_fullname}
\bibliography{references}

\end{document}